%% file: main.tex
\pdfoutput=1

\documentclass[11pt]{article}

\usepackage{acl}

\usepackage{times}
\usepackage{latexsym}

\usepackage[T1]{fontenc}

\usepackage[utf8]{inputenc}

\usepackage{microtype}
\usepackage{amsmath}
\usepackage{graphicx}

\newcommand{\mask}{\textit{<mask> }}
\newcommand{\rand}{\textit{<rand> }}

\newcommand{\name}{MANER }
\newcommand{\namens}{MANER}
\newcommand{\maskns}{\textit{<mask>}}
\newcommand{\randns}{\textit{<rand>}}
\newcommand\mc{\mathcal}
\newcommand\mb{\mathbf}
\renewcommand{\v}[1]{\mathbf{#1}} 

\title{MANER: Mask Augmented Named Entity Recognition\\
for Extreme Low-Resource Languages}

\author{Shashank Sonkar \\
  Rice University \\
  \texttt{ss164@rice.edu} \\\And
  Zichao Wang \\
  Rice University \\
  \texttt{jzwang@rice.edu} \\\And
  Richard G. Baraniuk \\
  Rice University \\
  \texttt{richb@rice.edu}
  }

\begin{document}
\maketitle
\begin{abstract}
This paper investigates the problem of Named Entity Recognition (NER) for extreme low-resource languages with only a few hundred tagged data samples.
NER is a fundamental task in Natural Language Processing (NLP).
A critical driver accelerating NER systems' progress is the existence of large-scale language corpora that enable NER systems to achieve outstanding performance in languages such as English and French with abundant training data. 
However, NER for low-resource languages remains relatively unexplored.
In this paper, we introduce Mask Augmented Named Entity Recognition (MANER), a new methodology that leverages the distributional hypothesis of pre-trained masked language models (MLMs) for NER. 
The \mask token in pre-trained MLMs encodes valuable semantic contextual information. 
MANER re-purposes the \mask token for NER prediction. 
Specifically, we prepend the \mask token to every word in a sentence for which we would like to predict the named entity tag.
During training, we jointly fine-tune the MLM and a new NER prediction head attached to each \mask token.
We demonstrate that MANER is well-suited for NER in low-resource languages; our experiments show that for 100 languages with as few as 100 training examples, it improves on state-of-the-art methods by up to $48\%$ and by $12\%$ on average on F1 score. 
We also perform detailed analyses and ablation studies to understand the scenarios that are best-suited to MANER.

\end{abstract}

\section{Introduction}
\input{sections/intro}

\section{Methodology}
\label{sec:method}
\input{sections/methodology}


\section{Experiment}
\label{sec:exp}
\input{sections/experiments}


\section{Conclusions}
\input{sections/conclusion}
In this paper, we have proposed Mask Augmented Named Entity Recognition (MANER) for NER in extreme low-resource language settings. 
MANER exploits the information encoded in pre-trained masked language models (inside \mask token specifically) and outperforms existing approaches for extreme low-resource languages with as few as only 100 training examples by up to $48\%$ and by $12\%$ on average on F1 score.
Analyses and ablation studies show that use of semantics encoded in \mask token is integral to MANER.
Future work will target using MANER for rapid prototyping in an active learning setup since it offers a cost-effective way to supplement human annotators. 




\bibliography{anthology,custom}
\bibliographystyle{acl_natbib}




\end{document}

%% file: sections/intro.tex

Named Entity Recognition (NER) is a fundamental problem in natural language processing (NLP)~\cite{Nadeau2007}.
Given an unstructured text, NER aims to label the named entity of each word, be it a person, a location, an organization, and so on.
NER is widely employed as an important first step in many NLP downstream applications, such as scientific information retrieval~\cite{Krallinger2005,Krallinger2017}, question answering~\cite{molla-etal-2006-named}, document classification~\cite{Guo2009}, and recommender systems~\cite{Jannach2022}.

Recent advances in NER have mainly been driven by deep learning based approaches, whose training relies heavily on large-scale datasets~\cite{scaling}. 
As a result, the most significant progress in NER is for resource-rich languages such as English~\cite{wang2020automated}, French~\cite{Tedeschi2021}, German~\cite{german}, and Chinese~\cite{Zhu2022}.
This reliance on large training datasets makes it challenging to apply deep learning-based NER approaches to low-resource languages where training data is scarce.
To illustrate the ubiquity of low-resource languages, WikiANN \cite{wikiann}, one of the largest NER datasets, has NER labeled data for 176 languages; however, 100 of those 176 languages have only 100 data samples.

Providing NER for low-resource languages is critical to ensure the equitable, fair, and democratized availability of NLP technologies that is required to achieve the goal of making such technologies universally available for all~\cite{low-resource-1,king2015practical}.
In addition to our team, several research teams are pushing the frontiers of NER for low-resource languages in two orthogonal and complementary directions.
The first direction aims to obtain larger NER datasets to solve the data scarcity problem, via either direct data collection or augmentation~\cite{malmasi-etal-2022-multiconer,annotate-ner,Meng2021,wikiann}. 
The second direction aims to develop new model architectures and training algorithms to account for the data scarcity. 
For example, ideas from meta-learning~\cite{deLichy2021}, distance supervision~\cite{meng2021distantly}, and transfer learning~\cite{transfer-learn-ner} leverage the few-shot generalizability of language models for NER in data-scarce settings.

\textbf{Contributions:}
In this work, we propose Mask Augmented Named Entity Recognition (MANER), a new approach for NER of low-resource languages that does not rely on additional data or modify existing model architectures.
MANER exploits the semantic information encoded in a pre-trained masked language model (MLM). 
Specifically, we reformat the input to the MLM by prepending a mask token to every token in the text to be annotated with NER tags.
This reformatted input is then used to fine-tune the MLM with a randomly initialized NER prediction head on top of the prepended mask tokens.

Extensive experiments on 100 extremely low-resource languages (each with only 100 training examples) 
demonstrate that MANER improves over state-of-the-art approaches by up to $48\%$ and by $12\%$ on average on F1 score.
Detailed ablation and analyses of MANER demonstrate the importance of using the encoded semantic information (\mask token vs.\ new random token, languages seen during pre-training vs.\ languages seen only during fine-training), and performance with respect to the number of training examples).

\begin{figure*}[t]
    \centering
    \includegraphics[scale=0.4]{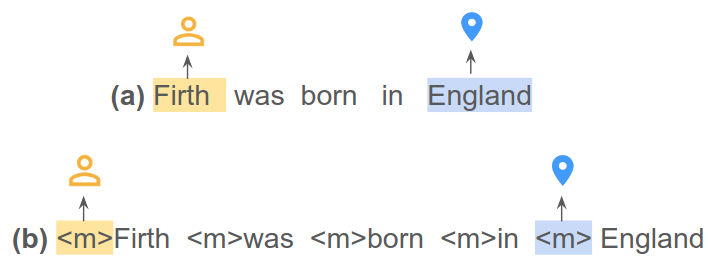}
    \caption{Proposed strategy to modify the input sentence using the \mask token which serves as input to our proposed 
    Mask Augmented Named Entity Recognition
    (\namens) model.}
    \label{fig:method}
\end{figure*}









%% file: sections/methodology.tex
In this section, we develop \name, which repurposes the \mask token for NER training. \name takes as input a sentence and outputs a NER label for each word in the sentence.

Let the set of NER labels be denoted by $\mc{N} = \{0, 1, 2, 3\}$, where $0={\sl N/A}$, $1={\sl PER}$, $2={\sl LOC}$, and $3={\sl ORG}$. 
Let the set of NER labels $L$ for a sentence $S$ consisting of $n$ words be $\{w_{0}, w_{1},.., w_{n-1}\}$ be $\{c_{0}, c_{1},.., c_{n-1}\}$ where $c_i \in \mc{N} \,, \ 0 \leq i<n$. 

In the following section, we first propose a key modification to the input sentence $S$ that incorporates the \mask token and second specify the design details of \namens.

\textbf{Proposed modification to input sentence:}
We \textit{append} a \mask token to beginning of each word in sentence $S$. The new sentence $S' = \{m, w_{0}, m, w_{1},.., m, w_{n-1}\}$ where $m$ is the \mask token.
The modified labels $L'$ are $\{c_{0}, \emptyset , c_{1}, \emptyset, .., c_{n-1}, \emptyset\}$. The original NER label of each word in the sentence is assigned to the \mask token to the immediate left of the word.

We hypothesize that in such a setting, \name will be better able to use the \mask token to weigh the relative relevance of the neighboring word vs.\ the rest of the context when determining the label to assign to the neighboring word.
For instance, \textit{Paris} is founder of a new AI startup called Banana.
Appending a \mask token before \textit{Paris} captures additional contextual information that may be helpful to classify the NER label of \textit{Paris}.

\textbf{MANER:} 
We now describe MANER, which consists of a transformer model and an NER classifier.
The transformer model takes a sentence as input and outputs embeddings for each token in the sentence.
The NER classifier uses the token embeddings to output the most probable NER class for each token. 

Denote the \name model by $\mc{M}$. The transformer model is given by $T$, $T(S') = T(\{m, w_{0}, m, w_{1},.., m, w_{n-1}\}) = \{\v{e}_{0}, \v{e}_{1},.., \v{e}_{2n-1}\}$, where $\v{e}_i \in \mb{R}^D \,, \ 0 \leq i<2n-1$, and $D$ is embedding dimension. 
The NER classifier is modeled using a weight matrix $\v{M} \in \mb{R}^{D \times |\mc{N}|}$ that takes the computed token embeddings as input. 
Using these token embeddings, the classifier output scores for all NER labels for each token in the sentence.
Passing these scores through a softmax nonlinearity provides probabilities $\v{p}_i \in \mb{R}^{|\mc{N}|}$ for all NER classes in $\mc{N}$ for a given token $i$ in $S$:
\begin{equation}
    \v{p}_i = \rm{softmax}\Bigl(\v{M}\bigl(\v{e}_i\bigl)\Bigl).
\end{equation}
Summing up, we have
\begin{equation}
    \mc{M}(T, \v{M}, S', i) = \v{p}_{i}, \, 0 \leq i <2n.
\end{equation}
The weights of $\v{M}$ and $T$ are learned/fine-tuned by minimizing cross-entropy loss.
Note that the loss is not calculated for labels marked $\emptyset$ in the modified label set $L'$. 
The NER label of the word is given by the NER label of the \mask token preceding it.

\textbf{\namens~inference:}
Similar to training, during inference, each token in the sentence is prepended with the \mask token and the NER class of each word is the most probable NER class of the \mask token prepended to it.

\begin{figure*}[t]
    \centering
    \includegraphics[width=\textwidth]{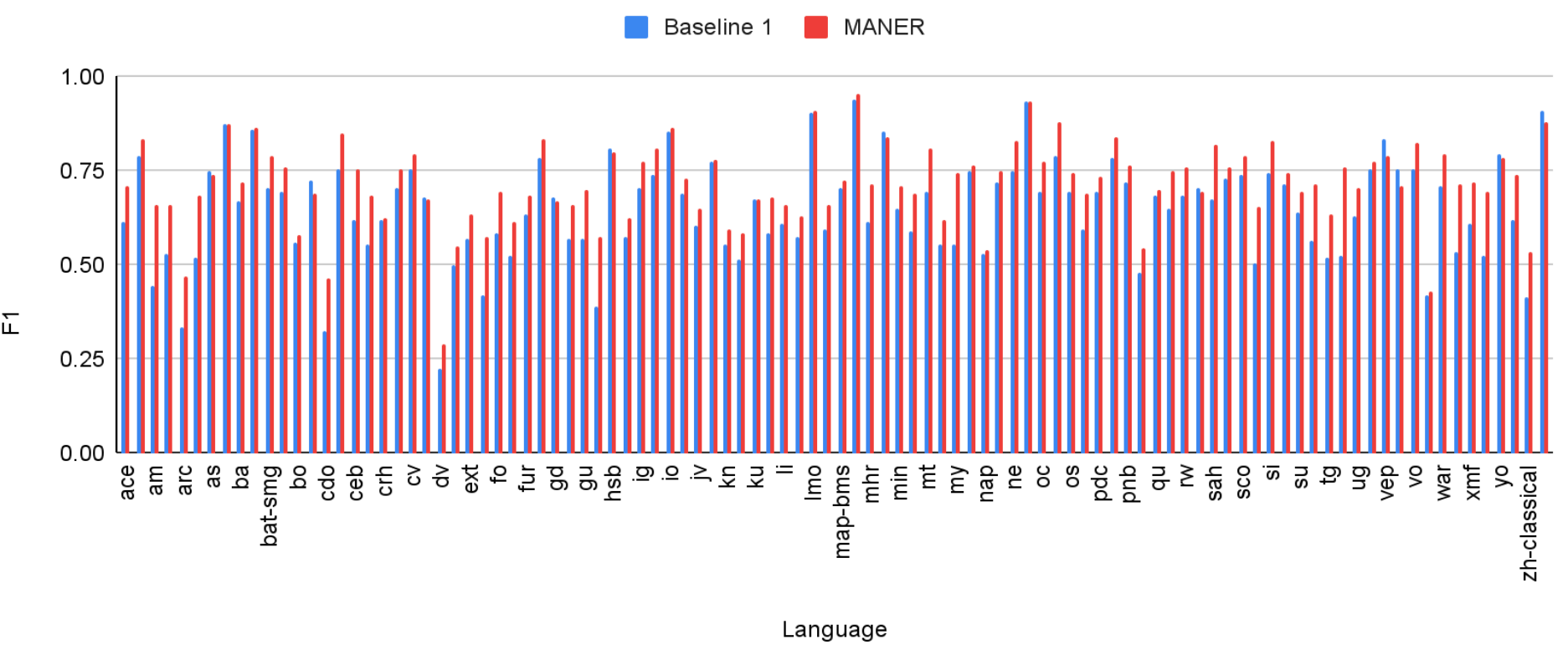}
    \caption{F1 scores comparing baseline NER methods against \name for 100 low-resource languages in the WikiANN dataset that only have 100 training samples each. On average, \name gives a significant improvement of \textbf{12\%} on F1 score compared to baseline as shown in Table \ref{tab:ner_main}.
    }
    \label{fig:100_strategy}
\end{figure*}

%% file: sections/experiments.tex

This section describes the NER dataset used for experiments and the quantitative results.
Then, we investigate why \name performs better for extreme low-resource languages using multiple qualitative measures.

\subsection{Dataset, Models, and Baselines}
\textbf{Dataset:} 
We use the WikiANN multilingual NER dataset \citep{panx,wikiann}, which provides named entities for Wikipedia articles across 176 languages.
To the best of our knowledge, WikiANN is by far the most comprehensive dataset for multilingual NER.
There are other multilingual datasets (e.g., CoNLL \cite{CoNLL2002,CoNLL2003}), but they cover only four popular languages:  English, German, Dutch, and Spanish.

Our main emphasis in this paper is on the 100 languages in  WikiANN that each have only \textbf{100 samples} for train and test splits.
The NER labels in WikiANN are in IOB2 (Inside–outside–beginning) format \cite{iob} comprising PER (person), LOC (location), and ORG (organization) tags.
An instance of NER tagged sentence: \textit{UNICEF}(B-ORG) is a nonprofit  organization, founded by \textit{Ludwik}(B-PER) \textit{Rajchman}(I-PER) headquartered at  \textit{New}(B-LOC) \textit{York}(I-LOC), \textit{United}(B-LOC) \textit{States}(I-LOC).

\textbf{Model:} We use XLM-RoBERTa-large \cite{xlm_roberta} as the transformer backend for \namens. XLM-RoBERTa-large is a multilingual version of the RoBERTa \cite{roberta} transformer model, designed for NER experiments on WikiANN.
XLM-RoBERTa model has been pre-trained using the MLM objective on 2.5TB of filtered CommonCrawl data containing 100 languages \cite{cc100}.
We chose this model, since it is the largest publicly available multilingual transformer based model.

\textbf{First Baseline:} 
Similar to \name design, current NER systems built upon transformer models also simply add a NER classifier to top of a transformer model.
The classifier predicts the NER class of each token of an unmodified sentence $S$:
\begin{align*}
        \mc{M}_{\rm base}(T, \v{M}, S, i) =  \rm{softmax}\Bigl(\v{M}\bigl(\v{e}_i\bigl)\Bigl) = \v{p}_{i}, \,
    \\ 0 \leq i <n,
\end{align*}
where $\mc{M}_{\rm base}$ is an NER model built on a transformer model $T$ using classifier weight matrix $\v{M}$.
This baseline method remains the de-facto method for training NER models for most languages (especially low-resource languages) to the best of our knowledge, though specialized models have been built for popular languages like English.

\textit{Inference:}
Similar to training during inference, the NER class of each word in the sentence is the most probable NER tag assigned to the classified word embedding.

\textbf{Second Baseline:}
Our MANER methodology in Section \ref{sec:method} is one way to change the input phrase using the mask token.
In this second baseline, we introduce yet another way to repurpose the \mask token for NER that is inspired by the masked language modeling (MLM) framework that is used for pre-training transformer models.
In MLM, a word is predicted using the words surrounding it in the sentence.
Since the NER category of a word is also a semantic property of the word, we use the philosophy of MLM for NER fine-tuning.

In MLM pre-training, the dataset is prepared by \textit{masking} random words in a sentence with a \mask token with a fixed probability $p_{mlm}$.
Then, the masked words are predicted using the context information.

Analogous to MLM pre-training, for NER fine-tuning, we randomly replace words in sentence $S$ with the \mask token with the fixed probability $p_{ner}$.
However, instead of predicting the missing words, as with MLM, we predict the NER labels $L$ for each word $w$ in $S$ irrespective of whether the word was replaced by a \mask token or not.
In the case the word was replaced with the \mask token, the transformer outputs the \mask token embedding for that word.

Thus the modified input to the transformer is $S' = \rm{mask}(S)$, where 
\begin{equation}
    \rm{mask}(w_i) = 
    \begin{cases}

        \text{\mask},& \text{if } p_i \leq p_{ner},\\
        w_i,              & \text{otherwise}
    \end{cases}
\end{equation}
with $p_i$ a random number between $0$ and $1$ generated for $w_i$.
Then, we use the first baseline NER model design $\mc{M}_{base}$ for training, but now it is is fine-tuned on $S'$ and $L$ (note we predict the label of the \mask tokens as well).
Inference with this model remains same as the first baseline model.

\subsection{Quantitative results and comparison of different strategies to repurpose \mask}

We train an XLM-RoBERTa-large model for 100 languages for 30 epochs with a learning rate of $5\mathrm{e}^{-6}$ and the loss optimized using Adam \cite{adamw}.
We report the average of F1
score for 100 languages in WikiANN that have 100 training samples each in Table \ref{tab:ner_main}.

\begin{table}[h]
\resizebox{\columnwidth}{!}{%
\begin{tabular}{|c|c|c|c|}
\hline
\textbf{Metric} & \textbf{Baseline 1} & \textbf{Baseline 2} & \textbf{\name} \\ \hline
F1 & 0.649 & 0.643 (-0.5\%) & 0.715 (12\%) \\ \hline
\end{tabular}
}
\caption{Average F1 scores for the 100 languages in the WikiANN dataset that have only 100 samples for the different strategies outlined in Figure \ref{fig:method}. In the low-resource language setting, MANER achieves a significant average improvement of \textbf{12\%}.}
\label{tab:ner_main}
\end{table}

\begin{figure*}[t]
    \centering
    \includegraphics[width=\linewidth]{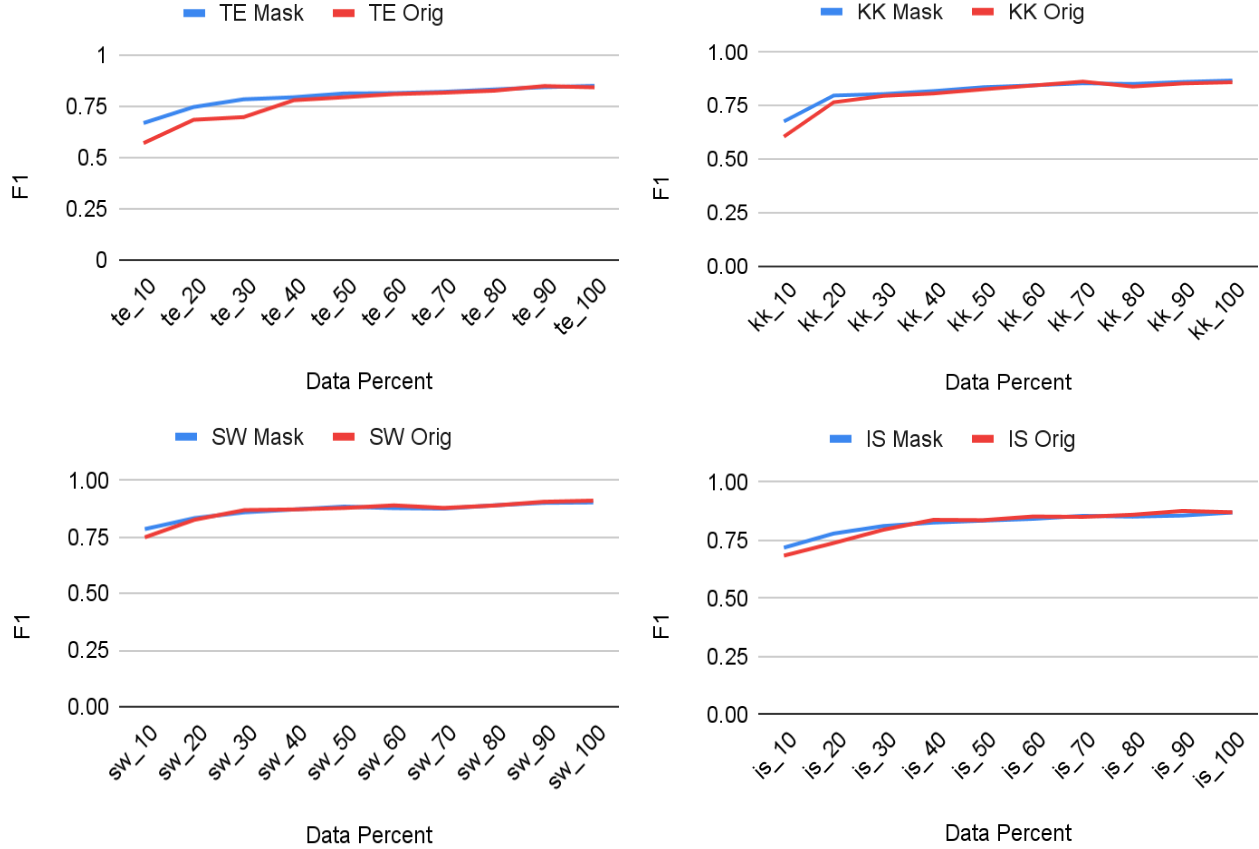}
    \caption{Measure effect of training samples to performance in \namens. \name can give a boost in performance till 400 samples and then both \name and first baseline NER model perform similarly. This demonstrates that \name is best suited for extreme low resource languages and rapid prototyping since it is easy and cost-effective to obtain very few human annotations to achieve large performance improvements (just 100 annotations are required).}
    \label{fig:sample}
\end{figure*}

MANER provides a significant \textbf{12\%} average improvement in F1 score in the low-resource language setting. 
Moreover, MLM based baseline 2 performs similarly to baseline 1.

Our interpretation for why the second baseline does not improve over \name (even though both methods use the \mask token), is primarily because of two reasons,
First, there is a difference between the train and inference setting in the second baseline.
In the inference setting of second baseline, the \mask token is not used to classify the token, unlike in \name.
\textit{In \name, the model can learn to give more importance to the context in the case of out-of-distribution test labels using the \mask token during inference and vice versa, explaining its significant 12\% improvement as compared to both baselines.}

The second reason is because the \mask token introduces noise in the context during training of the second baseline model. 
Hence, it loses critical information that makes the NER task more difficult for the model.
However, in \name, the \mask token is used in conjunction with the word to be labelled.

In Figure \ref{fig:100_strategy}, we plot the F1 score of all of the 100 low-resource languages comparing the first baseline against \name. One can see that there are only a few languages (12 out of 100) in which the first baseline outperforms \namens.

\subsection{Importance of the \mask token}

We now conduct experiments to demonstrate the importance of using the \mask token in \namens.
Intuitively the \mask token can be helpful because it encodes the semantics of the context and, thereby, the word (by distributional hypothesis) that needs to be tagged.

\textbf{Control token:} In the first experiment to gauge the importance of the \mask token, we replace the token with a control/random token (\randns) in \name.
Note that the \rand token is not learned during the XLM-RoBERTa model pre-training; thus it will not encode any contextual information.
As we see in Table \ref{tab:ner_control}, if we replace the \mask token with \rand in \namens, we achieve only a 6\% improvement in F1 performance over the first baseline.

\begin{table}[h]
\resizebox{\columnwidth}{!}{%
\begin{tabular}{|c|c|c|c|}
\hline
\textbf{Metric} & \textbf{Baseline 1} & \vtop{\hbox{\strut \textbf{\name}}\hbox{\strut \textbf{(w/ \maskns)}}} & \vtop{\hbox{\strut \textbf{\name}}\hbox{\strut \textbf{(w/ \randns)}}} \\ \hline
F1 & 0.649 & 0.715 (12\%) & 0.679 (6\%) \\ \hline
\end{tabular}
}
\caption{Average F1 scores for 100 languages for \name using the \mask token and the \rand token.
}
\label{tab:ner_control}
\end{table}

\textit{Why does \rand token provide a 6\% gain in F1?} We believe that \name provides a significant 6\% boost in F1 score even without using the \mask token because the model can use the \rand token to predict how much weight to assign the context and the word immediately adjacent to it depending on the test sample.

\textbf{XLM-RoBERTa pre-training languages:}
In the second experiment to gauge the importance of the \mask token, we report the F1 score in Table \ref{tab:ner_pretrain} averaging only on those languages that the XLM-RoBERTa model was pre-trained on with at least 0.5GB of training data per language.
The rationale behind this experiment design is that the \mask token will encode the context semantics of a language only if the language was seen during the pre-training stage of XLM-RoBERTa model.

As we see in Table \ref{tab:ner_pretrain}, in this case, \name provides a whooping 18\% improvement in F1 score (as compared to the 12\% gain in Table \ref{tab:ner_main}) if the language was seen in the pre-training stage.
This experiment again highlights the importance of using \mask token in \namens.

\begin{table}[h]
\centering
\resizebox{0.75\columnwidth}{!}{%
\begin{tabular}{|c|c|c|c|}
\hline
\textbf{Metric} & \textbf{Baseline 1} & \textbf{\name} \\ \hline
F1 & 0.603 & 0.705 (18\%) \\ \hline
\end{tabular}
}
\caption{
F1 scores comparing \name against the first baseline. F1 scores are averaged on languages that XLM-RoBERTa model was pre-trained on with at least 0.5GB of training data per language. The improvement over the baseline is 18\% compared to 12\% improvement in previous study that included all 100 languages.}
\label{tab:ner_pretrain}
\end{table}

\subsection{Effect of training samples}

In the previous subsections, we showed, by exploiting the context semantics encoded in the \mask token, \name can achieve significant gains over the baseline NER methodology for low-resource languages.
In this section, we measure the effectiveness of MANER in situations where more training data is available.
For this purpose, we select 4 languages from WikiANN dataset that have 1000 training data samples each.
From Figure \ref{fig:sample}, we see that \name boosts F1 performance over the the first baseline until about 400 samples and then both methods perform similarly. 
This demonstrates that \name is best suited for \textit{extreme low resource languages and rapid prototyping} since it is \textit{easy and cost-effective to obtain very few human annotations to achieve large performance improvements} (just 100 annotations are required).

The \textit{catastrophic forgetting} \cite{forget} phenomenon that masked language models undergo during any kind of fine-tuning is one of the reasons we think MANER does not provide gains when more training data is available (of course more training data also implies less reliance on specialized techniques like ours).
Catastrophic forgetting causes the loss of useful context semantics encoded in the \mask token during the fine-tuning stage that MANER heavily relies on.
Adding an additional masked language modeling loss to the NER loss during fine-tuning can help to circumvent catastrophic forgetting, but its investigation is left for future work.